\documentclass[conference]{IEEEtran}
\IEEEoverridecommandlockouts

\usepackage{cite}
\usepackage{amsmath,amssymb,amsfonts}
\usepackage{algorithmic}
\usepackage{graphicx}
\usepackage{textcomp}
\usepackage{xcolor}
\usepackage{multirow}
\usepackage{multicol}
\def\BibTeX{{\rm B\kern-.05em{\sc i\kern-.025em b}\kern-.08em
    T\kern-.1667em\lower.7ex\hbox{E}\kern-.125emX}}
\begin{document}

\title{VersaGauss: A Versatile Framework for Generating Multiphase Dynamics with 3D Gaussians}

\author{\IEEEauthorblockN{Ruijie Su\IEEEauthorrefmark{1},
Lingxiao Yang\IEEEauthorrefmark{2},
Xiaohua Xie\IEEEauthorrefmark{1}\IEEEauthorrefmark{3}, and
Jianhuang Lai\IEEEauthorrefmark{1}\IEEEauthorrefmark{3},}
\IEEEauthorblockA{\IEEEauthorrefmark{1}School of Computer Science and Engineering, Sun Yat-sen University, Guangzhou, China}
\IEEEauthorblockA{\IEEEauthorrefmark{2}School of Systems Science and Engineering, Sun Yat-sen University, Guangzhou, China}
\IEEEauthorblockA{\IEEEauthorrefmark{3}Guangdong Province Key Laboratory of Information Security Technology, Sun Yat-sen University, Guangzhou, China}
\thanks{
Corresponding author: Xiaohua Xie (email: xiexiaoh6@mail.sysu.edu.cn).

This work was supported by the Project of Guangdong Provincial Key Laboratory of Information Security Technology under Grant 2023B1212060026.}}

\maketitle

\begin{abstract}
Recent progress has been made in 3D Gaussian representation for reconstruction, generation, and physical simulation. However, current approaches mainly concentrate on physics-based dynamic generation of solid objects and only handle single-phase collision interactions. We introduce VersaGauss, a unified framework for generation, simulation, and rendering that supports versatile physics-based dynamic generation, particularly for multiphase interactions. Our system takes a few images as input and produces a realistic, physics-driven 3D dynamic scene with multiple objects. To optimize the Gaussian kernel distribution, we develop a particle pruning algorithm. We also propose the Coupled Multiphase Point Method (CMPM) to effectively model and generate multiphase interactions. Additionally, harmonic interpolation within CMPM and a Gaussian evolution strategy are introduced to achieve realistic fluid rendering. Extensive experiments demonstrate that our framework can simulate interactions among various materials such as fluid, rubber, sand, snow, and others. Code is available at https://github.com/Elowen-surj/VersaGauss.
\end{abstract}


\section{Introduction}
Recent advances in Neural Radiance Fields (NeRFs) \cite{mildenhall2021nerf} have enabled high-quality 3D reconstruction and novel-view synthesis. 3D Gaussian Splatting (3DGS) \cite{kerbl3Dgaussians} further improves efficiency by representing scenes with explicit 3D Gaussians and rendering via fast splatting, making it a popular choice for 3D generation and synthesis \cite{liang2024luciddreamer, yi2024gaussiandreamer}.

The explicit Gaussian representation also suggests a particle view: each Gaussian can be treated as a particle with physical attributes, allowing particle-based simulation to animate 3DGS scenes into physically plausible 4D content. This offers a principled alternative to purely generative 4D pipelines \cite{zheng2024unified, sun2024eg4d} and avoids visual artifacts introduced by multi-stage representation conversions in conventional simulation-to-rendering workflows.

\begin{figure}
	\centerline{\includegraphics[width=\linewidth]{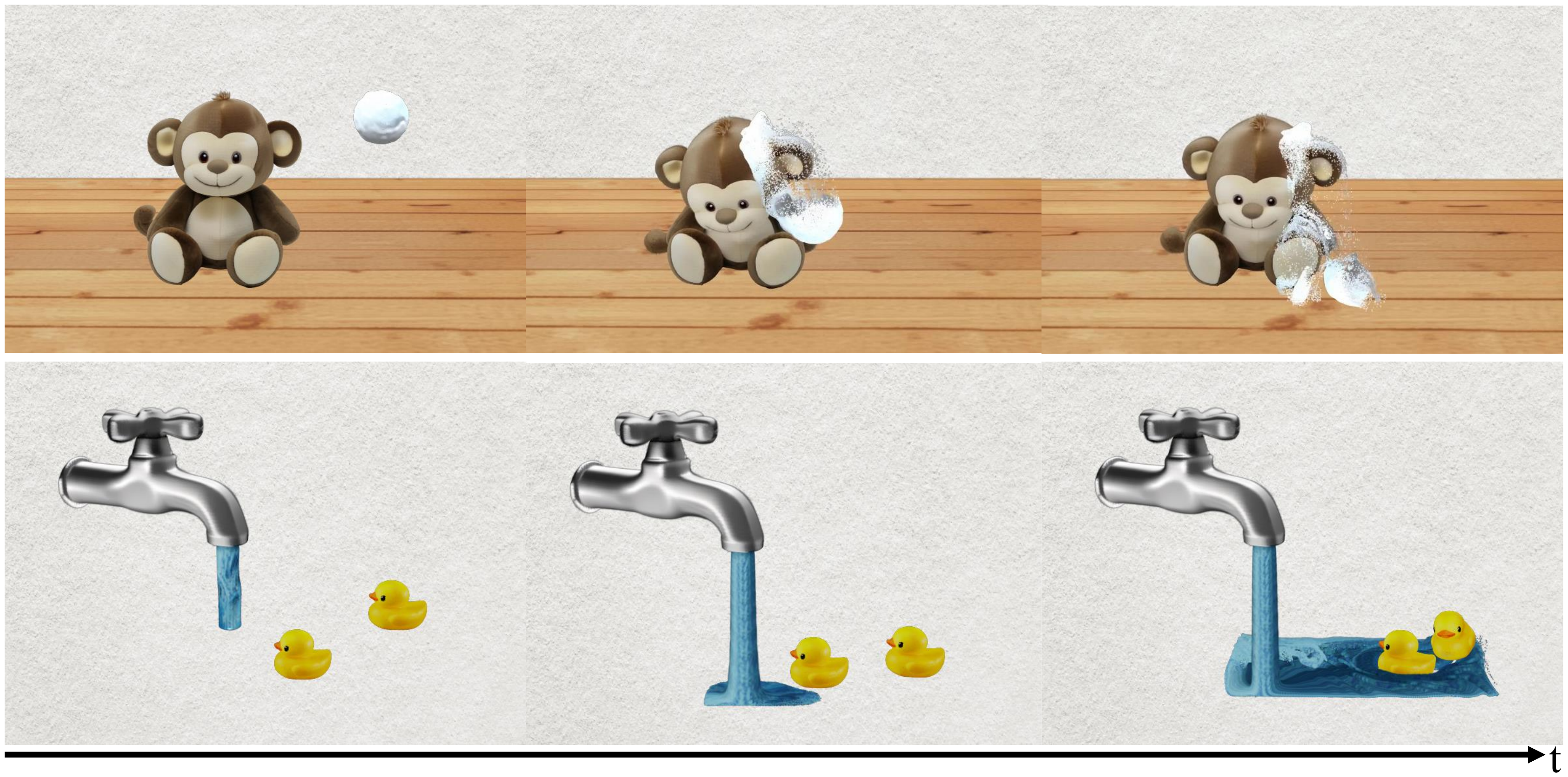}}
    \caption{VersaGauss is a novel, unified framework for generation, simulation, and rendering that achieves physics-based dynamics in multiphase interactions.}
	\label{fig:teaser} 
\end{figure}

Several works, such as PhysGaussian~\cite{xie2024physgaussian}, integrate 3D Gaussian Splatting (3DGS) with the Material Point Method (MPM) by assigning material parameters to Gaussian particles and deforming them according to simulated dynamics. Despite this progress, existing 3DGS-based physics methods remain limited in both material diversity and interaction scope: they typically support a limited set of phases (e.g., solids and some non-Newtonian fluids) and are typically limited to weak coupling within single-phase or same-material interactions. As a result, efficiently achieving accurate coupling dynamic among phases with disparate densities and constitutive behaviors is still difficult. Moreover, in fluid-involved scenarios, particles may freely cross object boundaries, leading to blur or noise.

We propose VersaGauss, a physics-based dynamics generation framework for coupled multi-phase interactions with realistic fluid appearance. Starting from a few images, VersaGauss segments objects \cite{kirillov2023segment}, reassembles them, and refines Gaussian distributions via particle pruning, while a particle–object–material linkage enables material-dependent behaviors. We further introduce Coupled Multiphase Point Method (CMPM) with material-specific constitutive models, together with harmonic interpolation and Gaussian evolution to stabilize and improve fluid rendering. VersaGauss outputs a dynamic 3D scene with multi-view rendering, generating a 50-frame video in about 10 minutes (comparable to recent video generators \cite{wan2025, yang2024cogvideox}).

Contributions. (1) We present VersaGauss, a unified generation-simulation-rendering framework for high-quality physics-based dynamics generation with multi-phase coupling in 3D GS. (2) We develop efficient strategies for multiphase interaction and fluid rendering, including particle pruning, harmonic interpolation, and Gaussian evolution, and introduce CMPM for coupled multiphase simulation. (3) We demonstrate versatile multiphase interactions through extensive experiments, showcasing competitive generation efficiency and scalability.

\section{Related Work}

\subsection{Large 3D Generation model}
Early 3D generation methods leveraged 2D diffusion priors through Score Distillation Sampling (SDS) \cite{yi2024gaussiandreamer, poole2022dreamfusion}. While capable of generating high-quality assets, these optimization-based approaches suffer from lengthy optimization times and geometric inconsistencies. A recent direction is to train feed-forward generators on large-scale 3D data for second-level synthesis. LRM \cite{hong2023lrm} predicts NeRFs \cite{mildenhall2021nerf} with a transformer, while follow-ups \cite{tang2024lgm} adopt Gaussian splatting and latent diffusion backbones to improve resolution and quality. TRELLIS \cite{xiang2024structured} proposes a unified 3D latent representation (SLAT) and rectified-flow transformers, enabling high-quality conditional 3D generation with flexible output representations. We use TRELLIS to obtain static 3D objects from masked images.

\subsection{Multiphysics Simulation Methods}
Multiphysics simulation targets coupled interactions among diverse materials (e.g., fluids, elastomers, plasticity). Existing methods broadly fall into (i) coupling specialized solvers and (ii) unified formulations. Coupling approaches \cite{xie2023contact} connect dedicated simulators (e.g., SPH \cite{ihmsen2014sph} for fluids, FEM for solids), offering flexibility but requiring complex interfaces and careful stabilization; weak coupling may introduce interface artifacts, while strong coupling is costly. Unified methods, such as PBD/XPBD \cite{macklin2016xpbd} and energy-based formulations \cite{li2020incremental}, provide tighter coupling and improved stability, but may trade accuracy across material classes and still face difficulties under extreme topology changes. 

Our simulation method, CMPM, is based on the Material Point Method (MPM), a unified particle-grid framework in which particles carry state information and the grid computes forces and updates momentum. MPM inherently supports large deformations and phase changes, and its grid projection facilitates momentum exchange between phases; additional contact models can be integrated as needed. Despite challenges like numerical viscosity and artificial adhesion, MPM offers a practical balance of stability, extensibility, and simplicity in implementation for our multiphysics context.

\subsection{Physics-based Dynamic Generation}

Recent work combines generative models with physical simulation for dynamic 3D/4D synthesis. Prior studies model motion via frequency/flow cues \cite{liu2024physgen} or embed simulators into NeRF pipelines \cite{abou2024particlenerf}. With 3DGS, PhysGaussian \cite{xie2024physgaussian} couples Newtonian dynamics to Gaussians via MPM, while Gaussian Splashing \cite{feng2024splashing} integrates PBD with 3D Gaussians for fluid simulation. Other efforts learn dynamic priors or adjust physical parameters guided by video generators \cite{huang2025dreamphysics}; Phys124 \cite{lin2024phy124} further scale physics-grounded 4D generation from images.

However, most existing methods handle a limited set of materials and primarily support interactions within a single phase, making complex multi-phase coupling challenging. Our goal is to generate coupled interactions across diverse phases, including fluids, rubber, sand, snow, and beyond.

\begin{figure*}[t!]
	\centering
	\includegraphics[width=\linewidth]{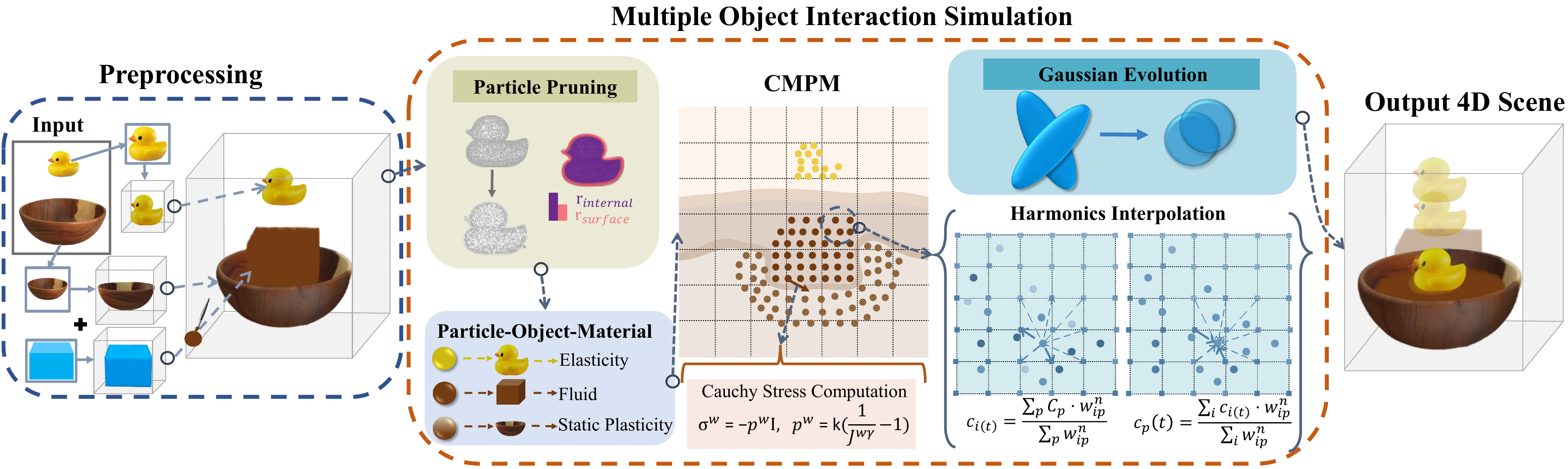}
	\caption{Method Overview. VersaGauss is a unified framework for generation, simulation, and rendering that facilitates the physical generation of interactions among objects composed of various phases with 3D Gaussians.}
        \label{fig:model}
\end{figure*}

\section{Preliminary Knowledge}

\noindent \textbf{3D Gaussian Splatting (3DGS).}
3DGS \cite{kerbl3Dgaussians} reconstructs and renders a scene using an explicit set of 3D Gaussians
$\{\mathbf{x}_p,\alpha_p,\mathbf{A}_p,\mathcal{C}_p\}_{p\in\mathcal{P}}$,
where $\mathbf{x}_p$ is the center, $\alpha_p$ the opacity, $\mathbf{A}_p$ the covariance, and $\mathcal{C}_p$ the spherical-harmonic (SH) color coefficients. Rendering projects each 3D Gaussian to a 2D Gaussian and the color at pixel $i$ is
\begin{equation}
  \mathcal{C}_i=\sum_{k} G_k(i)\alpha_k\,\mathrm{SH}(\mathbf{d}_k;\mathcal{C}_k)\prod_{j<k}\bigl(1-G_j(i)\alpha_j\bigr),
\end{equation}
where $G_k(i)$ is the 2D Gaussian weight and $\mathbf{d}_k$ is the view direction.

\noindent \textbf{Material Point Method (MPM).}
MPM is a particle-grid simulator: particles store material state (e.g., mass, velocity, deformation) and transfer it to a background grid to compute forces and update momentum; updated grid quantities are then transferred back to particles. This hybrid formulation is stable under large deformations and naturally handles contact and topological changes, making it well suited for complex material dynamics.

\section{VersaGauss}
As shown in Fig.~\ref{fig:model}, we present VersaGauss, a framework for physics-grounded dynamic 3D generation with Newtonian interactions across diverse phases. Given one or more images containing multiple objects, we segment each foreground object using SAM\cite{kirillov2023segment}. We then reconstruct each object with TRELLIS\cite{xiang2024structured} as a 3D Gaussian representation. To enable simulation, we interpret Gaussians as a particle cloud, rescale all objects into a canonical simulation domain, and reassemble them according to the original layout. For small objects, we apply \emph{particle pruning} to reduce particle count while preserving appearance. We also introduce a \emph{particle-object-material link} that assigns object-level material parameters to particles efficiently.

To simulate coupled multiphase interactions, VersaGauss integrates our Coupled Multiphase Point Method (CMPM). It supports weakly compressible fluids and phase-aware coupling by maintaining per-phase grid velocities. For realistic rendering of mixed fluids, we smooth surface appearance by transferring spherical-harmonic (SH) color attributes through the MPM grid and refine fluid Gaussian evolution (covariance/opacity) before rendering. The output is a dynamic 3D Gaussian scene that can be rendered from novel views.

\subsection{Preprocessing}

Current 3D generators are mainly trained for single-object reconstruction, which complicates multi-object dynamics. We first segment each object with SAM and crop it by the bounding box. Each crop is resized to a canonical, centered view and fed into TRELLIS\cite{xiang2024structured} to obtain a 3D Gaussian object.

We map all Gaussian centers into a simulation domain. Each object is normalized and then rescaled/repositioned to match its relative size and layout in the original image. When object $o_i$ is scaled by $s_{o_i}$, we adjust each Gaussian covariance as
\begin{equation}
\mathbf{A}'_p = \mathbf{A}_p \cdot s_{o_i}^{\frac{2}{3}}.
\end{equation}
After simulation, we apply the unified inverse transform for rendering.

\subsection{Multiple Object Interation Generation.}

\paragraph{\textbf{Particle Pruning.}}
As current 3D generators are mainly trained for single-object reconstruction, direct simulation is expensive since each object may contain hundreds of thousands of Gaussians. For small objects, such particle density is unnecessary. We therefore prune particles with different ratios for surface vs.\ interior to preserve surface structural integrity while reducing cost.

We voxelize each object with a grid resolution
\begin{equation}
l = \big(\mathcal{P}_{o_i}\, s'_{o_i}\big)^{\frac{1}{3}},\qquad s'_{o_i}\in[0,0.5],
\label{eq:pp_1}
\end{equation}
where $s'_{o_i}$ is the scale of the object $o_i$ and $\mathcal{P}_{o_i}$ is the particle count. We build an opacity grid from $\alpha_p$ and classify cells as surface/interior via ray intersection counts. We then apply pruning with the following ratios:
\begin{equation}
r_{\text{int}} = 1 - s'{o_i},\qquad r{\text{surf}} = 1 - (s'_{o_i})^{\frac{2}{3}}.
\end{equation}
These pruning ratios are empirically determined during experiments to balance computational efficiency and visual fidelity. To compensate for reduced density, we enlarge the remaining covariances:
\begin{equation}
\mathbf{A}^{\text{after}}_p=\mathbf{A}^{\text{before}}_p\cdot\left(\frac{1}{1-r}\right)^{\frac{2}{3}}.
\end{equation}
For moderately sized objects ($s'{o_i}\in[0.5,0.7]$), we increase surface pruning (e.g., $r{\text{surf}}=1-(s'_{o_i})^{\frac{1}{3}}$). Since TRELLIS Gaussians often concentrate near surfaces, we optionally fill interiors for larger objects to ensure accurate physical behavior \cite{xie2024physgaussian}.

\paragraph{\textbf{Particle-Object-Material Link}}
Particles within the same object typically share material parameters. Instead of storing per-particle properties, we link each particle to an object-level material entry (e.g., material type and Young's modulus), reducing memory and simplifying simulation. During MPM updates, through particle-object-material link, each particle queries its linked material to select the constitutive model and compute stresses, yielding material-specific motion and deformation.


\paragraph{\textbf{Coupled Multiphase Point Method}}
We simulate dynamics with proposed Coupled Multiphase Point Method (CMPM). For fluids, we use a weakly compressible fluid constitutive model\cite{tampubolon2017multi} with the corresponding partial Cauchy stress is defined as
\begin{equation}
  \boldsymbol{\sigma}^w=-p^w\mathbf{I},\qquad
  p^w=k\left(\frac{1}{(J^w)^{\gamma}}-1\right),
\end{equation}
where $k$ is the bulk modulus and $J^w=\det(\mathbf{F}^w)$ is the determinant of the deformation gradient of fluid particles. We track $J^w$ (instead of the full $\mathbf{F}^w$) and update it as
\begin{equation}
  J^{w,n+1}_p = \left(\mathbf{I}+\Delta t\cdot \mathrm{tr}(\nabla \mathbf{v}^{n+1}_{p,w})\right)J^{w,n}_p,
  \label{con:fluid_Jw_update}
\end{equation}
where $\mathbf{v}_{p, w}^{n+1}$ is the velocity of particle $p$ at time step $t^{n+1}$ and $\Delta{t}$ is the time step size.

Traditional MPM methods simulate multi-phase interactions by computing grid node velocities as weighted averages of all nearby particles, disregarding phase-specific interaction dynamics. This blending causes particles near interfaces to share similar velocities, failing to capture complex inter-phase coupling.

To address this, we explicitly maintain separate velocities for each phase at every grid node, allowing co-located particles of different phases to retain distinct velocities and enabling phase-specific velocity updates. Thanks to the parallel simulation algorithm, this operation does not noticeably increase computation time.

Given the distinct interaction behaviors, we adopt separate update rules for fluid-fluid and fluid-sand mixing and integrate them into the CMPM. Following \cite{ren2014multiple}, fluid-fluid velocity updates at grid node $i$ for material $k$ are:

\begin{equation}
  {v}_{ik}^{n+1} = v^n_{im} + (\frac{\boldsymbol{f}^{ext}_i - a^n_{ik}\boldsymbol{f}^n_{im}}{m^n_{ik}})\Delta{t} + \tau(\frac{\nabla a_{ik}^n}{a_{ik}^n} - \sum_{k'}{c^n_{ik'}\frac{\nabla a_{ik'}^n}{a_{ik'}^n}}),
  \label{con:fluid-mixture}
\end{equation}
where $a^n_{ik} = \frac{V_{ik}^n}{V_{im}^n}$ and $c^n_{ik} = \frac{m_{ik}^n}{m_{im}^n}$ denote volume and mass fractions of fluid $k$, and $\tau$ is the diffusion coefficient. The terms represent the prior mixed velocity, stress and external forces, and diffusion velocity, respectively.

For fluid-sand interactions, following \cite{tampubolon2017multi}, the update at grid node $i$ for phase $k$ is:
\begin{equation}
  v^{n+1}_{ik} = v^{n}_{ik} + (\frac{\boldsymbol{f}^{ext}_i - \boldsymbol{f}^n_{ik}}{m^n_{ik}})\Delta{t} - \frac{c_E m_{ik} m_{ik'}(v_{ik}^{n+1} - v_{ik'}^{n+1}))}{m_{ik}^n},
  \label{con: fluid and sand}
\end{equation}
where $c_E = \frac{n^2\rho^w g}{p_E}$ is the drag coefficient dependent on sand porosity $n$, permeability $p_E$, and gravity $g$; $\boldsymbol{f}^n_{ik}$ is the phase-specific stress; $k'$ denotes the coupled phase (sand or fluid).

To enhance fluid surface fidelity and avoid artificial blobbiness, we integrate a kernel-weighted interpolation of spherical harmonics (SH) color coefficients into the CMPM, inspired by SPH \cite{ihmsen2014sph}. We extend grid nodes with color attributes and update them concurrently with other material properties such as mass and velocity:
\begin{equation}
  c_i(t) = \frac{\sum_p{\mathcal{C}_p}\cdot{w^n_{ip}}}{\sum_p{{w^n_{ip}}}},
\end{equation}
where $\nabla{w^n_{ip}}$ is the quadratic B-spline kernel evaluated at the position $x^n_p$. The updated colors are then transferred back to particles during the grid-to-particle step via:
\begin{equation}
  c_p(t) = \frac{\sum_i{c_i(t)}\cdot w^n_{ip}}{\sum_i{ w^n_{ip}}}.
\end{equation}

In summary, representing the 3D scene using Gaussian-based particles with assigned material properties enables multi-phase simulation with phase-specific constitutive models and incorporates updated rules to simulate specialized interaction coupling. Coupled with SH-based color interpolation for realistic visual effects, this approach yields physically plausible interactions and smooth fluid surface rendering.

\subsection{Rendering}
After updating Gaussian kernel positions via simulation, we render the deformed 3D Gaussian scene. Unlike solids, fluid particles exhibit large movements and may oscillate across object boundaries, often causing blurry artifacts or noisy renderings. Moreover, most 3D reconstruction models lack support for semi-transparency, and the high opacity of fluid particles can lead to oversaturated colors during dynamic events like splashes.

To address these issues, we propose a pre-rendering adjustment of fluid particle covariance and opacity. Before simulation, we voxelize the object region and estimate particle volume from voxel size and particle count. To avoid visual distortion from inconsistent volumes, we normalize particle volumes throughout the simulation. Using these volumes and particle radii, we update fluid particle covariance matrices to prevent artifacts from extreme values. We further refine covariance based on deformation.

As shown in Eq.(\ref{con:fluid_Jw_update}), fluid particles track the determinant of the deformation gradient $J_w$ rather than the full gradient. Thus, fluid deformation updates differ from solids.

We approximate the fluid kernel deformation via a first-order mapping applied to each kernel:
\begin{equation}
  \widetilde{\phi}(\boldsymbol{X}, t) = x_p + (J^{w}_p)^{\frac{1}{3}}\rm{I}(\boldsymbol{X} - \boldsymbol{X}_p),
\end{equation}
where $J^w$ is the determinant of the deformation gradient. The Gaussian distribution transforms as:
\begin{equation}
  G^w_p(x, t) = e^{-\frac{1}{2}(x-x_p)^T\left((J^{w}_p)^{\frac{1}{3}}\rm{I}\,\cdot\,\bar{\boldsymbol{A}}\,\cdot\, {(J^{w}_p)}^{\frac{1}{3}}\rm{I}\right)^{-1}(x - x_p)}.
\end{equation}
This yields a time-dependent 3D Gaussian Splatting (3DGS) formulation:
\begin{equation}
  x_p(t) = \widetilde{\phi}(\boldsymbol{X}, t),\quad \boldsymbol{a}_p(t) = (J^{w}_p)^{\frac{2}{3}}\,\cdot\,\rm{I}\,\cdot\,\bar{\boldsymbol{A}},
\end{equation}
scaling fluid particles isotropically to maintain spherical shape and minimize artifacts.

To model fluid semi-transparency, we introduce an opacity scaling factor $s_{\text{opacity}, o_i}$ for each fluid Gaussian kernel $p$ of object $o_i$, adjusting opacity as $\alpha_p' = \alpha_p \cdot s_{\text{opacity}, o_i}$, where $s_{\text{opacity}, o_i} \in [0.005, 0.02]$. Since interior regions are densely filled, this adjustment preserves overall appearance while improving splash realism and reducing noise.

In summary, given static 3D fluid Gaussians $\left\{\boldsymbol{X}_p, \boldsymbol{A}_p, \alpha_p, \mathcal{C}_p\right\}$, we produce dynamic Gaussians $\left\{x_p(t), \boldsymbol{a}_p(t), \alpha_p', c_p(t)\right\}$ over simulation time for rendering.

Finally, with the processed Gaussian scene and camera pose information, we synthesize physically plausible fluid motion videos via 3D Gaussian Splatting, providing coherent visualizations of the simulation.

\section{Experiments}

\begin{figure}[]
  \centering
	\includegraphics[width=\linewidth]{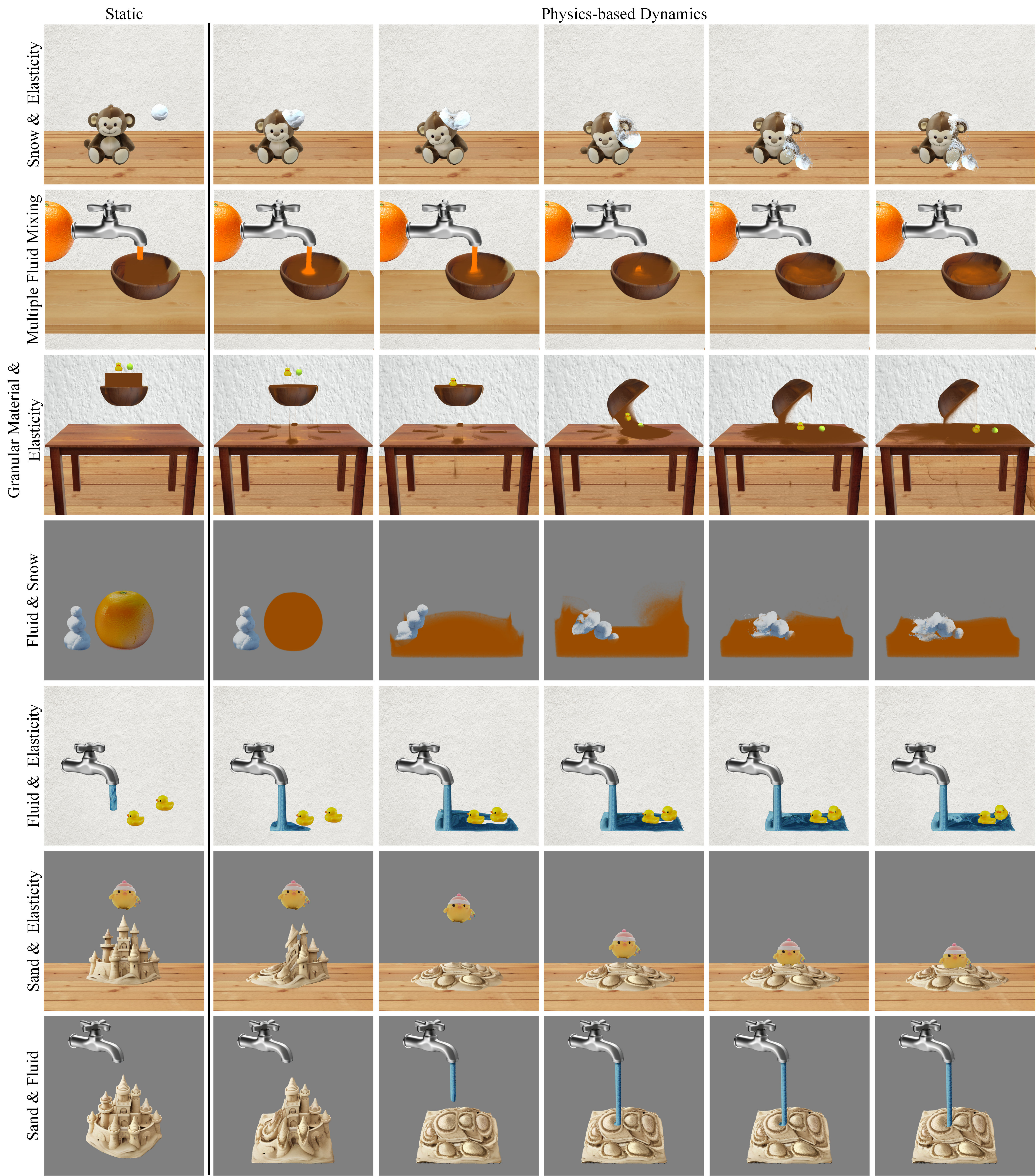}
	\caption{Generation Results. We presents 3D dynamic generation results generated by our VersaGauss framework, showcasing the interactions between various phases.}
        \label{fig: result2}
\end{figure}

\subsection{Results}

\paragraph{\textbf{Implementation Details}} Our CMPM simulator is built on the Warp framework \cite{macklin2022warp}, a Python library optimized for high-performance GPU simulation with parallel computing support. Simulations use a gravitational acceleration of $9.8 m/s^2$. We selectively freeze velocities of certain objects to induce controlled motion, while others follow physically natural dynamics. All experiments run on a single NVIDIA A100 GPU.

\paragraph{\textbf{Showcases}} Fig.~\ref{fig: result2} illustrates VersaGauss’s versatility in simulating multiphase interactions across diverse scenarios:
\textbf{1) Snow and Elastic Objects}: A snowball impacts a monkey doll, causing elastic deformation of the doll and fracturing of the snowball.
\textbf{2) Fluid-Fluid Mixing}: Fluids with differing densities and diffusion velocities gradually mix.
\textbf{3) Granular Material and Elastic Objects}: Elastic objects float or sink in granular flow depending on density; tilting the container causes them to move along with the flow.
\textbf{4) Fluid and Snow}: Snow fractures under fluid force and floats due to lower density.
\textbf{5) Fluid and Elastic Objects}: Elastic objects are carried by water flow.
\textbf{6) Elastic Objects and Sand}: A sandcastle collapses, and then elastic objects previously frozen fall into the mound, forming a depression.
\textbf{7) Fluid and Sand}: The sandcastle collapses, and the sand sinks with fluid flow under viscosity and gravity.

\begin{figure}[h]
  \centering
	\includegraphics[width=\linewidth]{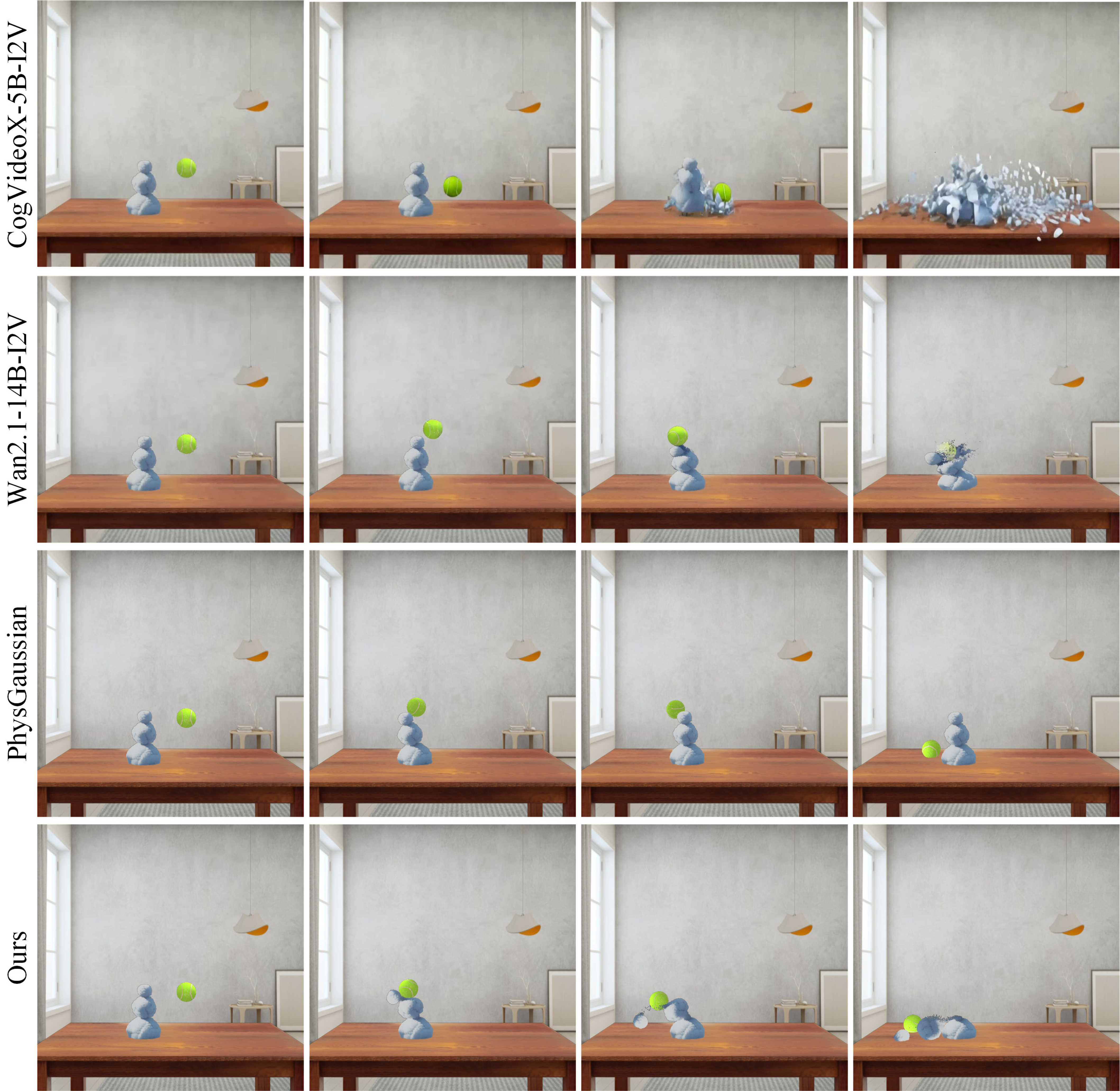}
	\caption{Comparison of ball–snowman dynamics. Given an initial image and prompt describing the physical motion, we generate videos using Wan2.1 \cite{wan2025} and CogVideoX-5B \cite{yang2024cogvideox}, and compare them with PhysGaussian \cite{xie2024physgaussian} and our method to evaluate realism in multiphase dynamics.}
        \label{fig: comparison}
\end{figure}

\paragraph{\textbf{Comparison}}
Fig.~\ref{fig: comparison} compares ball–snowman dynamic generation results produced by our method, PhysGaussian \cite{xie2024physgaussian}, and video generation models (Wan2.1, CogVideoX-5B). Our approach achieves physically accurate multiphase dynamics, whereas PhysGaussian and the video models fail to capture such interactions.

\begin{figure}[t]
  \centering
	\includegraphics[width=\linewidth]{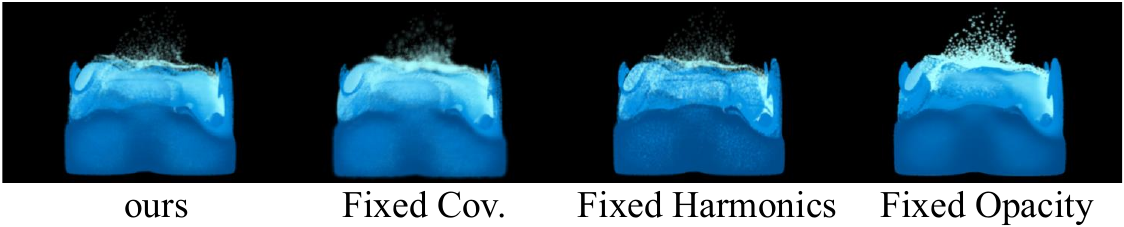}
	\caption{Ablation studies of Gaussian evolution and harmonics interpolation.}
        \label{fig: ablation}
\end{figure}

\begin{figure}[h]
  \centering
	\includegraphics[width=\linewidth]{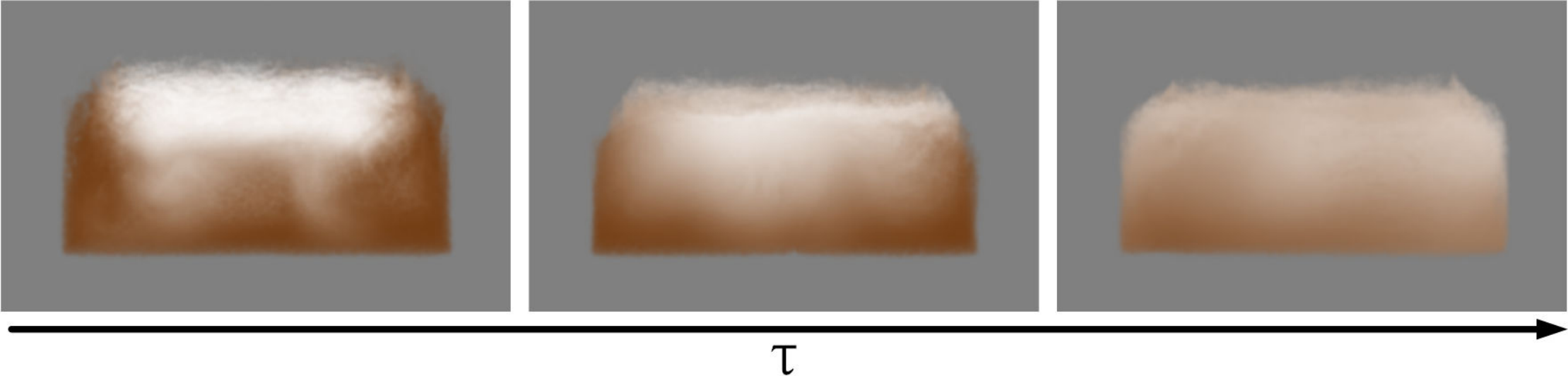}
	\caption{Fluid–Fluid Mixing. Mixing behavior of two fluid phases with different diffusion coefficients~$\tau$.}
        \label{fig: fluid_fluid}
\end{figure}
\begin{figure}[h]
  \centering
	\includegraphics[width=\linewidth]{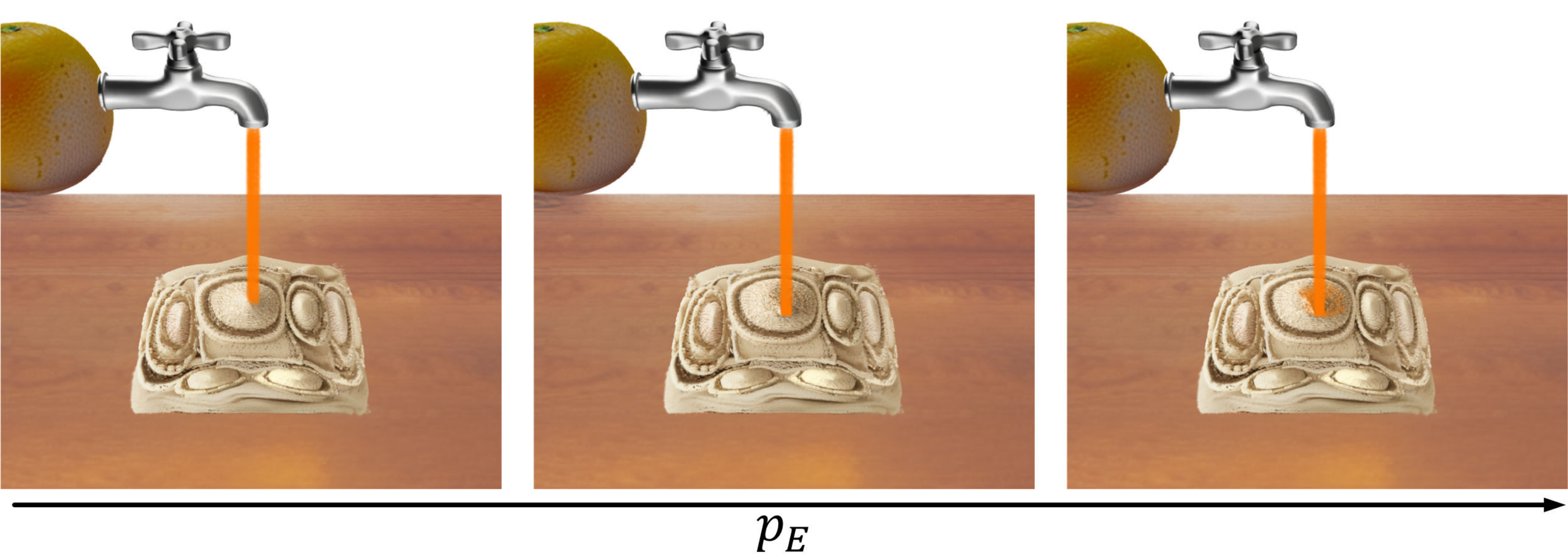}
	\caption{Fluid–Sand Interaction. Sand sinking 2 seconds after fluid contact under different sand permeabilities~$p_E$.}
        \label{fig: fluid_sand}
\end{figure}

\subsection{Ablation Studies}
We perform ablation studies to validate the effects of our pre-rendering adjustments on fluid Gaussian kernels: \textbf{(1) Fixed Covariance} uses the initial covariance throughout simulation and rendering; \textbf{(2) Fixed Harmonics} skips interpolation of color spherical harmonics coefficients; \textbf{(3) Fixed Opacity} omits opacity scaling. As shown in Fig.~\ref{fig: ablation}, fixed covariance causes blurry artifacts due to inappropriate kernel sizing, fixed harmonics produce sandy fluid surfaces, and fixed opacity leads to oversaturated renderings.

To evaluate the effects of covariance normalization by particle radii and deformation (via the deformation gradient determinant), we isolate these operations in rendering fluid splashes in Appendix~A.

Table~\ref{tab:my-table} reports statistics on Gaussian kernel count, simulation time, and rendering time with and without particle pruning. PSNR comparisons confirm that the proposed particle pruning method significantly reduces computational cost while preserving image quality.

\begin{table}[h]
\centering
\caption{We report the number of Gaussian kernels, simulation time, rendering time, and PSNR with and without particle pruning.}
\label{tab:my-table}
\resizebox{\columnwidth}{!}{%
\begin{tabular}{lccccl}
\hline
Scene & Pruning & Kernels ($10^{3}$) & \shortstack{Sim. Time \\ per Frame (s)} & \shortstack{Render Time \\ per Frame (s)} & PSNR \\ \hline
\multirow{2}{*}{\begin{tabular}[c]{@{}l@{}}balls \& duck \\ fluid\end{tabular}} & no & 368 & 25.48 & 0.26 & \multirow{2}{*}{45.79} \\
 & yes & 191 & 8.78 & 0.13 &  \\ \hline
\multirow{2}{*}{\begin{tabular}[c]{@{}l@{}}faucet \& ducks\end{tabular}} & no & 367 & 15.18 & 0.18 & \multirow{2}{*}{47.26} \\
 & yes & 71 & 4.84 & 0.07 &  \\ \hline
\end{tabular}%
}
\end{table}

\subsection{Additional Studies}

\paragraph{\textbf{Fluid-Fluid Mixing}}
Fig.~\ref{fig: fluid_fluid} illustrates our framework’s capability to simulate mixing between two fluid phases with different diffusion coefficients~$\tau$. The density ratio is set to white:brown = 1:2. As indicated by the third term in Eq.~(\ref{con:fluid-mixture}), $\tau$ primarily controls the diffusion velocity: low $\tau$ leads to stratification due to density contrast, while higher $\tau$ facilitates gradual mixing.

\paragraph{\textbf{Fluid-Sand Mixing}}
Fig.~\ref{fig: fluid_sand} shows the effect of sand permeability $p_E$ on fluid-sand interaction. Increasing $p_E$ enhances viscous coupling from sand particles, causing more sand to sink with the fluid flow.

Details on the influence of the fluid bulk modulus~$k$, the energy stability of CMPM, and experiments on automatic physical-parameter optimization using a video diffusion model with SDS loss are provided in Appendix~B.

\section{Discussion}

\paragraph{\textbf{Conclusion}}
VersaGauss presents a unified framework for generation, simulation, and rendering, enabling high-quality, realistic dynamics of multiphase interactions within a 3D Gaussian representation.

\paragraph{\textbf{Limitations}}
Current material types are primarily manually specified. Future work will integrate large language models and video generation models to automatically infer object materials and parameters in complex scenes.

\bibliographystyle{IEEEbib}
\bibliography{icme2026references}

\end{document}